\documentclass[runningheads]{llncs}
\usepackage[T1]{fontenc}

\usepackage{graphicx}
\usepackage[round]{natbib}
\usepackage{bm}
\usepackage{url}
\usepackage{xspace}
\usepackage{amsmath,amsfonts,amssymb,amscd,amsthm,xspace,mathtools}
\usepackage[hidelinks]{hyperref}
\usepackage{xcolor}
\usepackage[ruled,vlined]{algorithm2e}
\usepackage{comment}
\usepackage{algorithmic}
\usepackage{booktabs}
\usepackage{enumitem}

\begin{document}

\title{Power Grid Control with Graph-Based Distributed Reinforcement Learning}

\author{
Carlo Fabrizio\inst{1,*}
\and
Gianvito Losapio\inst{1,*}
\and
Marco Mussi\inst{1}
\and
Alberto Maria Metelli\inst{1} 
\and
Marcello Restelli\inst{1}
}

\authorrunning{Fabrizio et al.}

\institute{\textsuperscript{1}Politecnico di Milano, Milan, Italy}

\maketitle            

\begin{abstract}
The necessary integration of renewable energy sources, combined with the expanding scale of power networks, presents significant challenges in controlling modern power grids. Traditional control systems, which are human and optimization-based, struggle to adapt and to scale in such an evolving context, motivating the exploration of more dynamic and distributed control strategies. This work advances a \emph{graph-based distributed reinforcement learning} framework for real-time, scalable grid management. The proposed architecture consists of a network of distributed low-level agents acting on individual power lines and coordinated by a high-level manager agent. A Graph Neural Network (GNN) is employed to encode the network's topological information within the single low-level agent's observation. To accelerate convergence and enhance learning stability, the framework integrates imitation learning and potential-based reward shaping. In contrast to conventional decentralized approaches that decompose only the action space while relying on global observations, this method also decomposes the observation space. Each low-level agent acts based on a structured and informative local view of the environment constructed through the GNN. Experiments on the Grid2Op simulation environment show the effectiveness of the approach, which consistently outperforms the standard baseline commonly adopted in the field. Additionally, the proposed model proves to be much more computationally efficient than the simulation-based Expert method.
\end{abstract}

\section{Introduction}
\label{sec:introduction}
Power grids across the world are controlled from control centers by \textit{human dispatchers}, who monitor the electricity network 24 hours per day, 365 days per year. They must constantly keep the network within its thermal limits, frequency ranges, and voltage ranges by taking \textit{remedial actions} on network elements such as lines and substations via remote control command \citep{kelly2020reinforcement}.

Modern power grids are undergoing a strong transformation due to the introduction of renewable energy sources. Unlike conventional non-renewable generators, renewable energy sources (such as solar and wind) introduce high variability and limited dispatchability in the energy production since they depend on the atmospheric conditions.

Traditional power grid control systems, strongly reliant on human-in-the-loop and optimization-based algorithms, are unable to cope with the increasing complexity of such control problems. In order to overcome this complexity, it is required to develop new methods. Among emerging techniques, Reinforcement Learning (RL) has been explored in several works over the last years with promising results \citep{dorfer2022power, chauhan2023powrl, vandersar2023multiagentreinforcementlearningpower, lehna2024hugo}.

In order to investigate the potential use of RL methods within next-generation power grid controllers, the French electricity network management company RTE (R\'eseau de Transport d'\'Electricit\'e) developed a series of challenges called “Learning to Run a Power Network” \citep[L2RPN,][]{marot2021learningrunpowernetwork}. These challenges were designed to simulate realistic sequential decision-making environments, in which agents must maintain the power grid operative under different conditions of uncertainty. The challenges were offered over the Grid2Op framework, developed by RTE. Specifically, the Grid2Op framework provides realistic simulations of power grids and supports a wide range of control actions and interventions \citep{grid2op}.

The main advantage of using RL is that it can identify and also capitalize on under-utilized, cost-effective actions that human dispatchers and traditional solution techniques are unaware of or unaccustomed to, resulting in more efficient control of the power grid by learning effective relations in a data-driven way.

In parallel, the growing size of power grids, i.e., the increasing number of controllable elements and complex interconnections, results in a combinatorial explosion of potential control configurations. With this large state and action spaces, RL algorithms suffer a problem known as the \textit{curse of dimensionality}, i.e., the amount of data/computation required to achieve a good solution may be out of reach. Distributed Reinforcement Learning (DRL) algorithms can be considered to mitigate this problem by distributing the learning process among multiple agents~\citep{zhang2021multi}. 

\paragraph{Original Contribution.}
In this work, we propose a DRL algorithm for power grid control. In our solution, each agent uses a state representation based on a Graph Neural Network \citep[GNN,][]{wu2020comprehensive}. GNNs are a class of neural networks designed to work directly with graph-structured data. In a graph, data is represented as nodes (entities) and edges (relationships), which makes GNNs well-suited for problems where the structure or connections between elements are important, such as power grids. The combination of RL and GNNs allows to learn a control strategy that captures dependencies and interactions between connected nodes and efficiently handles large-scale graphs using message passing, where nodes iteratively aggregate information from their neighbors. This work focuses on a specific class of control actions named topology-based control, which are particularly relevant in power grid operations due to their zero operational cost. These actions operate by altering the electrical connectivity of controllable components within the grid. Unlike interventions on generators -- which incur in financial costs -- topology reconfigurations offer a cost-free alternative for redirecting power flows and maintaining grid stability. However, the complexity of such actions lies in their discrete, combinatorial nature and their non-local effects on the grid. 

This work proposes a scalable distributed graph-based RL architecture for large-scale power grids. The main contributions can be summarized as follows:

\begin{itemize}[itemsep=1pt]
    \item \textbf{Graph-Based Representation:} Development of an effective homogeneous graph representation of the power grid, enabling informative encoding of the system's topology.
    \item \textbf{Modular Control Architecture:} Design and implementation of a two-layer distributed architecture, consisting of low-level agents acting on power lines and a high-level controller coordinating their behavior, promoting modularity and scalability.
    \item \textbf{Observation Space Decomposition:} Integration of GNNs to perform feature extraction and implicitly decompose the observation space, allowing agents to operate on partial but informative views of the grid.
    \item \textbf{Imitation Learning for Accelerated Training:} Implementation of Deep Q-Learning from Demonstrations \citep[DQfD,][]{hester2017deepqlearningdemonstrations} to inject expert behavior into the training process, significantly accelerating convergence.
    \item \textbf{Reward Decomposition:} Design of a bootstrapped potential-based reward shaping strategy \citep{adamczyk2025bootstrappedrewardshaping} aimed at improving credit assignment and learning efficiency, particularly in large-scale power grid scenarios.
\end{itemize}

\paragraph{Paper Structure.}
The paper is organized as follows. Section~\ref{sec:problemformulation} provides the problem formulation and discusses the learning objective. Section~\ref{sec:proposedsolution} contains a detailed description of our proposed solution. Section~\ref{sec:experiments} shows the experimental results. Section~\ref{sec:relatedworks} presents related works, while Section~\ref{sec:conclusions} concludes the paper with possible future work.

\section{Problem Formulation}
\label{sec:problemformulation}
The work was conducted over the Grid2Op simulation environment, which is designed to simulate realistic power grids for a given period, usually several days, at 5-minute intervals. At each timestep, the agent is called to take an action on the simulator, leading to the next simulator’s state. The power grid is represented as a graph of connected elements: nodes and edges represent, respectively, substations and power lines. The simulator also models other elements, specifically, generators and loads.  Generators produce electricity while loads consume it. Substations can be viewed as routers within the power network: they encompass two internal buses to which their elements (loads, generators, or power lines) can be connected. Substations have no other role than controlling their internal connections, affecting the overall flow of energy. 
Among the various interventions supported by the simulator, topology-based actions are undoubtedly the most interesting and extensively studied. This is because they have no operational cost and are highly complex due to their combinatorial nature, making them particularly challenging for humans to execute effectively.  These actions allow modifications to the internal connectivity within substations. Thus, a substation $i$ controlling $N_i$ elements can perform up to $2^{N_i}$ actions, corresponding to all possible configurations of its elements across the two buses.
To ensure system stability, Grid2Op allows only one substation to be controlled at each time step, as simultaneous interventions may cause unpredictable interactions and instabilities.
The simulation runs for at most $T$ time steps, but it ends prematurely in case of a blackout. A blackout can occur if the energy demand (load requests) is not satisfied by the current configuration. The primary cause of a blackout is known as “line overload”. When the current flow of a specific power line exceeds a certain threshold, the simulator automatically disconnects the power line. This can potentially lead to a grid configuration that violates the load requirements. Grid2Op offers a large number of features at each time step to describe the environment's state, more details can be found in \citep{grid2op}.
The objective of a control algorithm is to keep the power grid stable despite the unknown evolution of energy production and consumption. As a performance metric, we consider the so-called “survive time”, which corresponds to the number of time steps the grid remains operational before a blackout occurs during an episode. The goal of an algorithm in this scenario is to maximize the survive time.

Notably, as the size of the power grid grows, the action space grows with the number of controllable elements. Moreover, the state representation becomes hard to be learned effectively due to the expanding grid’s graph structure and increasing number of complex interconnections.
An efficient control algorithm must therefore handle an extremely large action space while leveraging graph-structured data to generate informative observations.

\section{Proposed Solution}
\label{sec:proposedsolution}
In this section, we present the proposed solution, detailing the key components and procedures. First, an overview of the learning components is provided, followed by a description of the procedure used to convert the power grid state into a homogeneous graph.

\subsection{Distributed RL Module}
The core idea of the proposed solution is to design a two-layer distributed model that decomposes the large combinatorial action space into subsets, allowing each to be managed almost independently.
Rather than relying on a full global view of the system, the approach uses local observations. To make these observations more informative, the model leverages graph-structured data, allowing agents to extract meaningful features from their local neighborhood and better cope with the partial observability inherent in such distributed settings. The proposed control system is composed of:

\begin{enumerate}[itemsep=1pt]
\item \textbf{Low-Level Agents}: Deep Dueling Double Q-learning agents with Prioritized Experience Replay buffer \citep{wang2016dueling} in charge of managing the power lines. Each agent controls a distinct power line, performing only the topology actions associated with that specific line. Moreover, each agent perceives only its power line, receiving a local observation composed exclusively by the subset of features related to that power line. Therefore, from a single agent's perspective, all the other agents belong to the environment, which becomes complex and partially observable. Although common reinforcement learning multi-agent approaches for power grid management assign agents to substations \citep{yoon2021winning,vandersar2023multiagentreinforcementlearningpower,demol2025centrallycoordinatedmultiagentreinforcement}, we chose to assign agents to power lines instead, which more naturally aligns with the objective of controlling the grid stability, avoiding lines' overloads.
\item \textbf{Shared GNN}: all low-level agents share a common GNN, which processes a graph-based representation of the power grid to enrich the information available to the single agents. The GNN acts as a feature extractor that enhances the informativeness of the single line observation by incorporating neighborhood information. The Graph Neural Network aims to reduce the degree of partial observability of the environment from each low-level agent's perspective. Moreover, since the GNN is shared among agents, it implicitly promotes cooperation and possibly enhances generalization capabilities. This procedure potentially allows graph-neural transfer learning, where a model trained on a smaller grid can be used to initialize the model on a larger one, possibly improving convergence.
\item \textbf{High-Level Controller}: an RL controller that manages the agents, deciding which agent has to act on a specific situation. It receives as input the power lines' thermal limits and current flow information, and a binary topology vector. In a dangerous situation, the controller selects which agent has to act. Differently from many of the distributed solutions, which involve a rule-based selection procedure called CAPA \citep{yoon2021winning,vandersar2023multiagentreinforcementlearningpower}, this work enables a more flexible solution by means of an RL top-level agent, following a new promising research direction as \citep{manczak2023hierarchicalreinforcementlearningpower,demol2025centrallycoordinatedmultiagentreinforcement}. In this case, the high-level controller is a Deep Dueling Double Q-learning agent with Prioritized Experience Replay Buffer \citep{wang2016dueling}.
\end{enumerate}

Since convergence in such a two-layer Multi-Agent setting is hard to achieve, both low-level and top-level agents are pre-trained to learn expert demonstrations deriving from an Expert simulation-based algorithm \citep{marot2018expert}, by means of Deep-Q-Learning from demonstrations \citep{hester2017deepqlearningdemonstrations}. Moreover, in such a multi-agent collaborative setting, the use of a global reward may lead to instabilities or difficulties during learning, therefore, a potential-based reward-shaping technique \citep{adamczyk2025bootstrappedrewardshaping} is implemented, yielding interesting results.

\begin{figure}[t!]
    \centering
    \includegraphics[width=\textwidth]{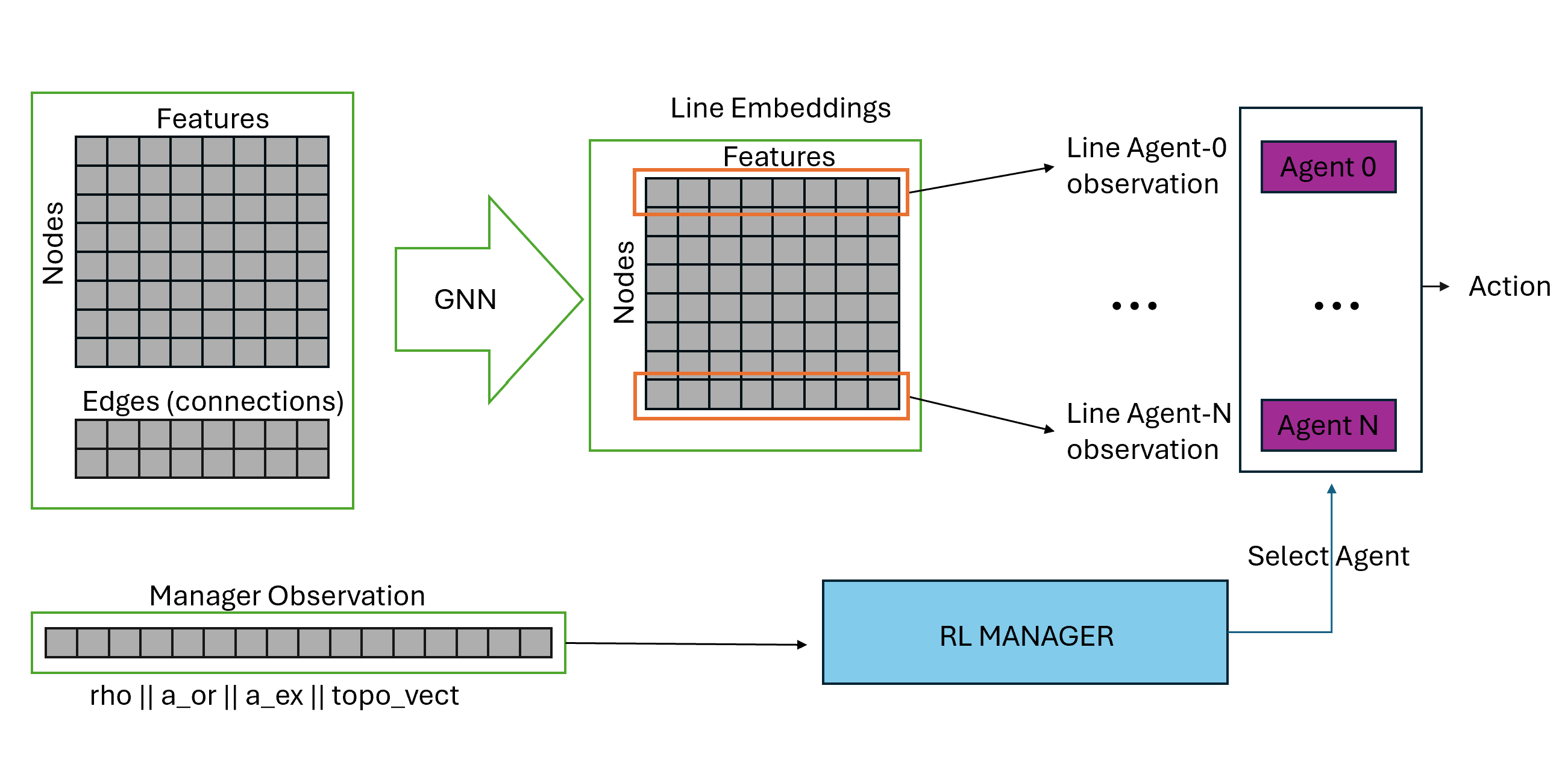}
    \caption{Abstract high-level representation of the model.}
    \label{fig: hlmod}
\end{figure}

To better illustrate the computational steps required to produce an action, Alg. \ref{alg:comp_flow} outlines the flow executed by the architecture when called to act on the environment. An abstract high-level representation of the model is provided in Fig. \ref{fig: hlmod}.

\begin{algorithm}[t!]
\caption{Action computation flow.}
\label{alg:comp_flow}
\begin{algorithmic}[1]
\STATE $i \leftarrow$ Manager.select\_agent$(s^t|\mu^t)$
\STATE $g^t \leftarrow$ convert\_graph$(s^t)$ 
\STATE  $o^t \leftarrow$ GNN$(g^t | \phi^t)[i]$
\STATE $a^t \leftarrow$ Agent$_i$.policy\_play$(o^t | \theta_i^t,\alpha_i^t, \beta_i^t)$
\STATE \textbf{return} $a^t$
\end{algorithmic}
\end{algorithm}

\subsection{Graph-Based Observation Construction}

The raw simulator observation has been pre-processed to obtain graph-like data before passing it to the Graph Neural Network. Both the Grid2Op simulator and the research community have proposed methods for representing the power grid as a graph; however, each approach presents certain limitations. All of the existing methods work by considering each element as a node (e.g., load, generator, power line, bus), raising the problem of graph heterogeneity: nodes present dissimilar features or belong to different classes. In general, the problem of graph heterogeneity strongly affects the performance of GNNs \citep{zhu2020homophilygraphneuralnetworks}, which usually fails to generalize in that setting. Therefore, the common approach to handling heterogeneous graphs with Graph Neural Networks is to employ models specifically designed for such structures, which inevitably increases model complexity \citep{li2023hetero2netheterophilyawarerepresentationlearning,zhu2020homophilygraphneuralnetworks,mao2023hinormerrepresentationlearningheterogeneous}.

Moreover, most of the existing methods represent buses in the graph but omit substations, meaning that the two buses belonging to the same substation are not connected. This leads to what is known as “bus-bar information asymmetry”, where each bus lacks access to information about potential connections originating from its counterpart within the same substation. The preprocessing procedure of this work is novel, returns a homogeneous graph representation and naturally handles the problem of busbar information asymmetry without using heterogeneous graphs as in \citep{dejong2025generalizablegraphneuralnetworks}.  

Initially, every element of the grid, except for substations, is directly represented as a vector. Then, the vectors of the elements connected to the same bus are summed to create the bus embedding. To represent a substation, the embeddings of its buses, including the bus of disconnected elements, are concatenated to form the substation embeddings. Finally, the power lines' embeddings are obtained by concatenating the embeddings of their two terminal substations.

\begin{figure}[t!]
    \centering
    \includegraphics[width=0.65\textwidth]{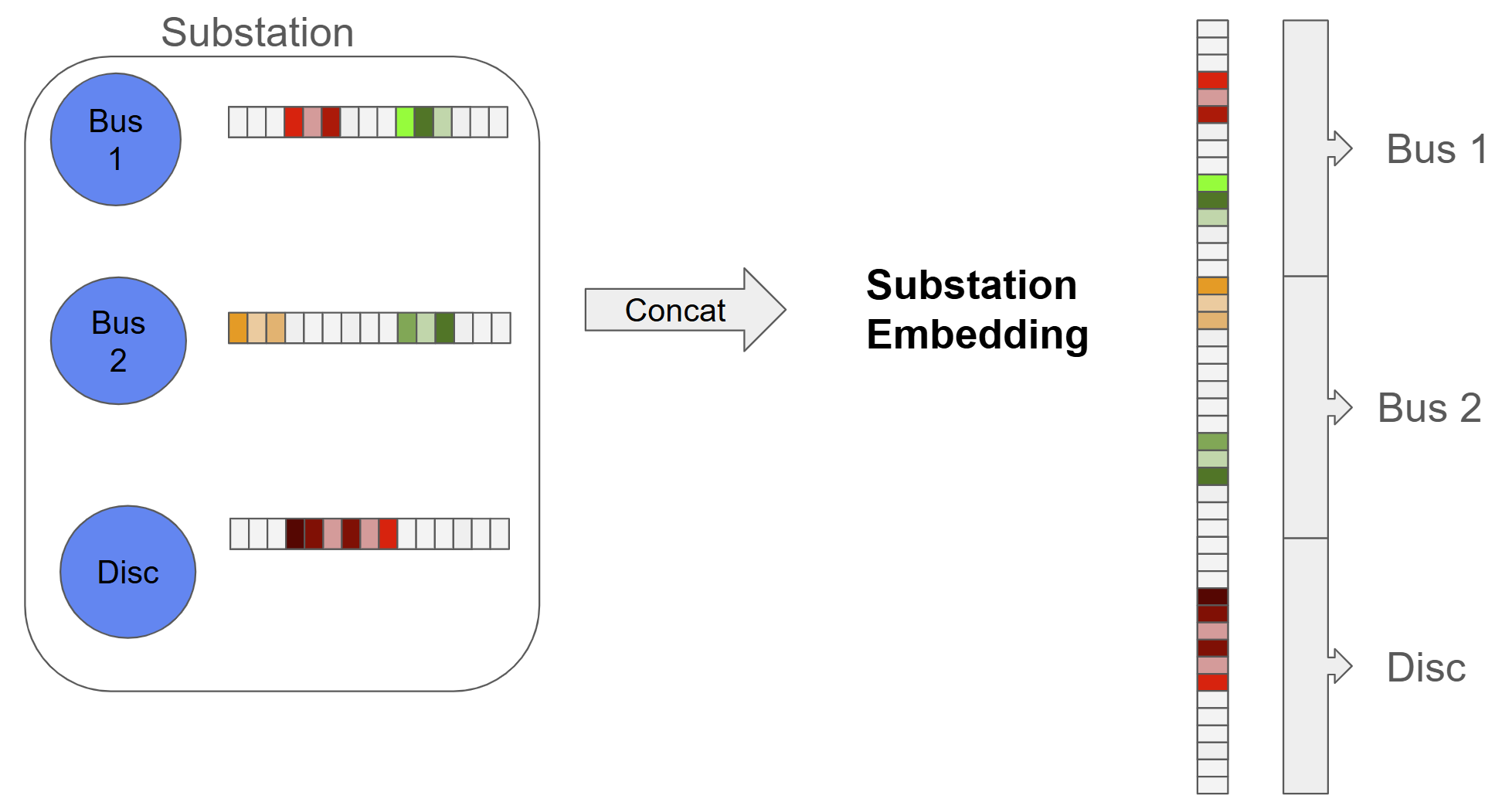}
    \caption{Example of substations embedding.}
    \label{fig: subemb}
\end{figure}

\begin{figure}[t!]
    \centering
    \includegraphics[width=0.65\textwidth]{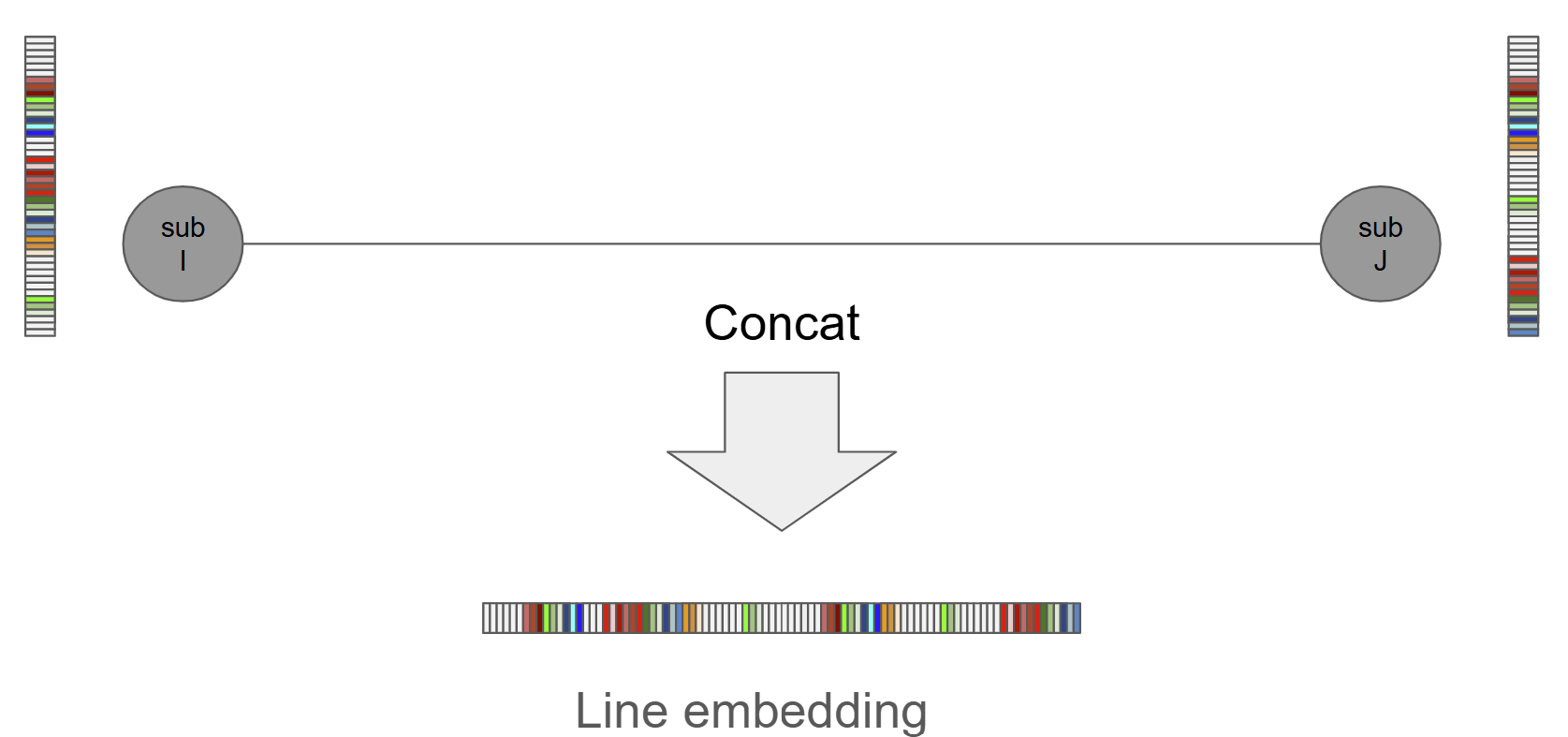}
    \caption{Example of lines embedding.}
    \label{fig: linembe}
\end{figure}

Each element (power line, load, generator, or bus) is represented as a feature vector using a one-hot-like encoding, where the vector is zero in all positions except for the subset of dimensions reserved for its specific type. In this way, it is possible to represent a substation by considering the vectors of the elements connected to it. By summing the vectors bus-wise, we obtain the bus embeddings, which are then concatenated to construct the substation embeddings, as shown in Fig. \ref{fig: subemb}. Finally, the line embedding is created as the concatenation of the embeddings of its origin and extremity substations as illustrated in Fig. \ref{fig: linembe}.

At this point, each line is represented by a vector that captures the information from both substations it connects. To build the graph, each power line is treated as a node, and two nodes are connected if their corresponding power lines share a common substation. This transformation converts the original power grid into a line graph, where power lines (originally edges) become nodes, and substations (originally nodes) become edges. With this procedure, a sequence of power grid observations is turned into a sequence of undirected homogeneous graphs. 

\section{Experiments}
\label{sec:experiments}

In this section, we empirically validate our model on the Grid2Op environment. The code used to run the experiments presented in this section is available at: \url{https://github.com/Carlo000ml/G4DQN}.

\subsection{Environment Description}
\begin{figure}[t!]
    \centering
    \includegraphics[width=\textwidth]{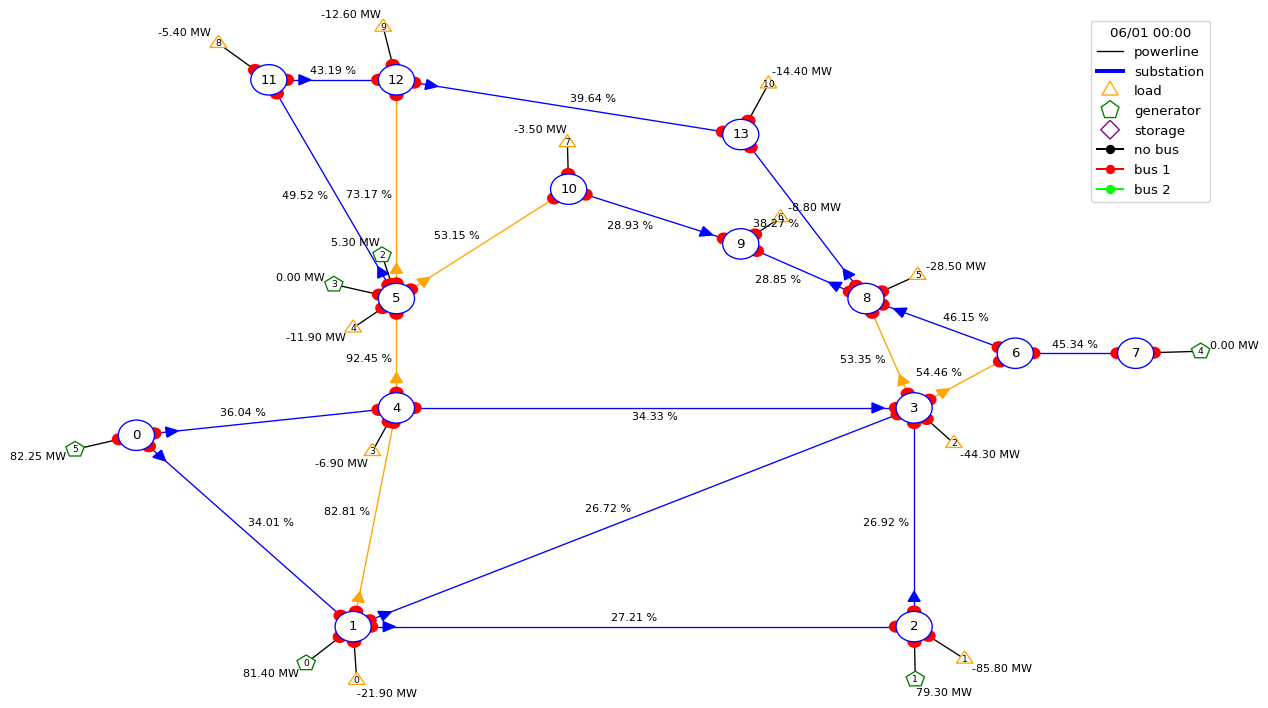}
    \caption{Representation of the Grid2Op \textit{“l2rpn\_case14\_sandbox”} power grid.}
    \label{fig: sandbox}
\end{figure}

The analysis was conducted on the \textit{“l2rpn\_case14\_sandbox”} environment of the Grid2Op simulator (Fig. \ref{fig: sandbox}). It represents a power grid counting 14 substations, 20 lines, 6 generators, and 11 loads, and it encompasses 1004 different episodes. Each episode is represented as a set of time series (referred to as a chronics), which models the behavior of generators and loads over time. The agent must keep the power grid stable despite the unknown evolution of energy production (time series on generators) and consumption (time series on loads). The simulation runs for a maximum of 8064 time steps. If the agent survives until the end, the simulation terminates regardless of whether a blackout has occurred. 

\subsection{Training Configuration}
Each power line is controlled by a dedicated low-level agent, resulting in 20 such agents operating in parallel. Including the RL manager, the system consists of 21 learning agents. Each agent maintains a pair of neural networks -- a main network and a target network. A GNN is employed as a feature extractor, with both main and target versions. Overall, 44 neural networks are simultaneously loaded on the computing device, with 22 actively trained and the remaining 22 kept frozen for target computation. Carefully chosen strategies were necessary to ensure stable and effective learning.
To this end, gradient clipping and soft main-target synchronization were adopted to achieve stable learning and effective training.
Each main neural network was optimized using the ADAM optimizer \citep{kingma2017adammethodstochasticoptimization}.
Inspired by the original DQN work \citep{DQN}, gradients were clipped in the range $[-3,3]$. In addition, soft updates between the main and target networks were employed, as this is a widely adopted practice for enhancing training stability in Deep Q-learning methods \citep{lillicrap2019continuouscontroldeepreinforcement}. Nevertheless, the most significant contribution to training stability came from increasing the synchronization interval. Synchronizing the main and target networks too frequently was indeed observed to cause an exponentially growing loss.
The most straightforward and easily configurable $\epsilon$ decay strategy based on half-life was selected for all experiments. In this approach, the value of $\epsilon$ decays according to an exponential schedule defined by a half-life parameter, which represents the number of steps required for 
$\epsilon$ to be reduced by half \citep{pytorchDQN}.

The proposed architecture is evaluated and tested against the \textit{Do-Nothing} baseline.
This baseline represents a passive control strategy where the agent takes no action throughout the episode, regardless of the system’s state. The strength of this policy lies in the design of the environment itself. By design, the initial configuration, where all components in the grid are connected to bus 1 of their respective substations, is a stable configuration. This setup allows the system to maintain safe operation for a considerable duration without intervention, especially under moderate or favorable chronics. As a result, the Do-Nothing policy often achieves non-trivial survival times and serves as a meaningful reference for evaluating learned strategies.

\subsection{Results}

\begin{figure}[t!]
    \centering
    \includegraphics[width=\textwidth]{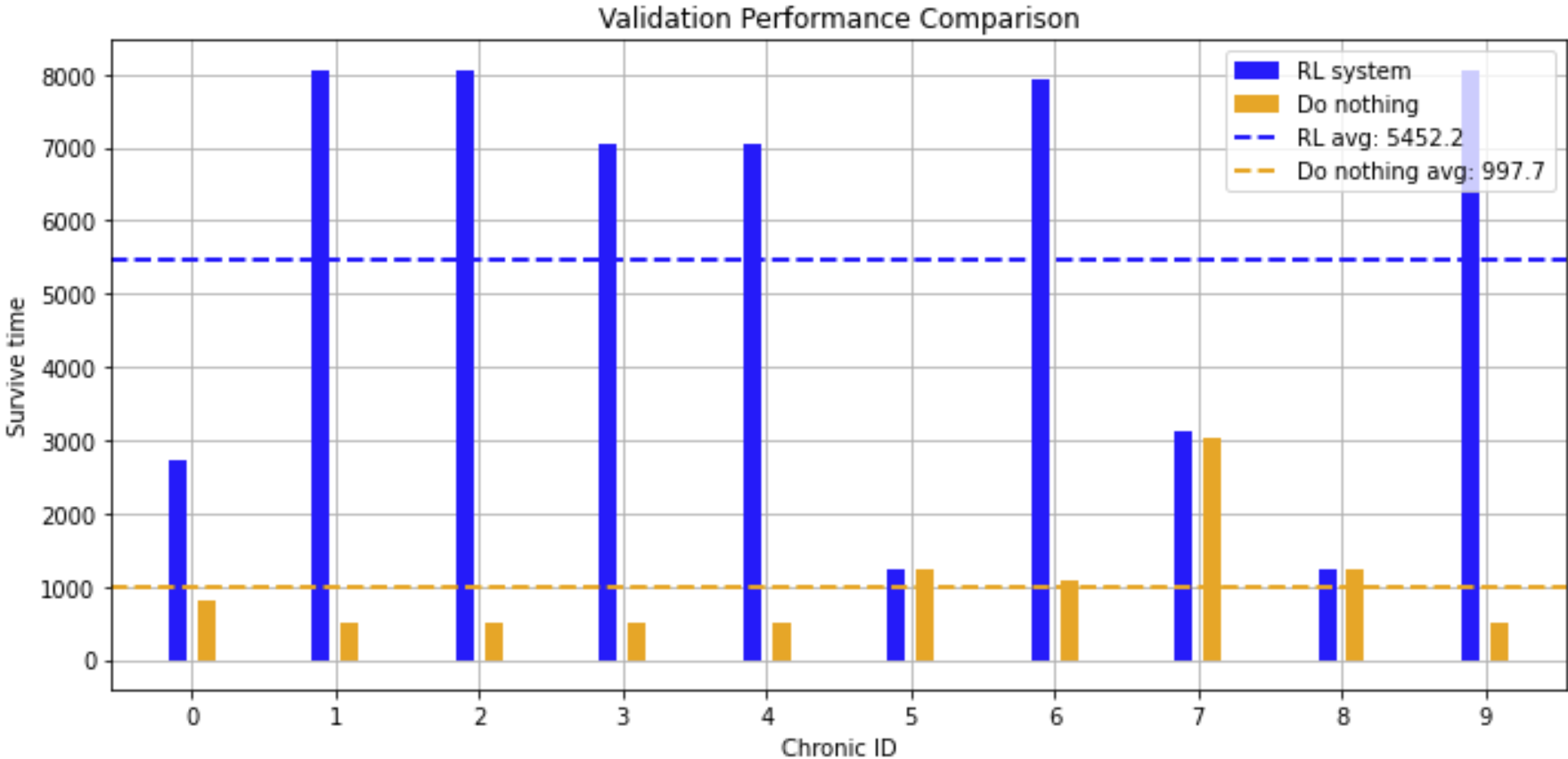}
    \includegraphics[width=\textwidth]{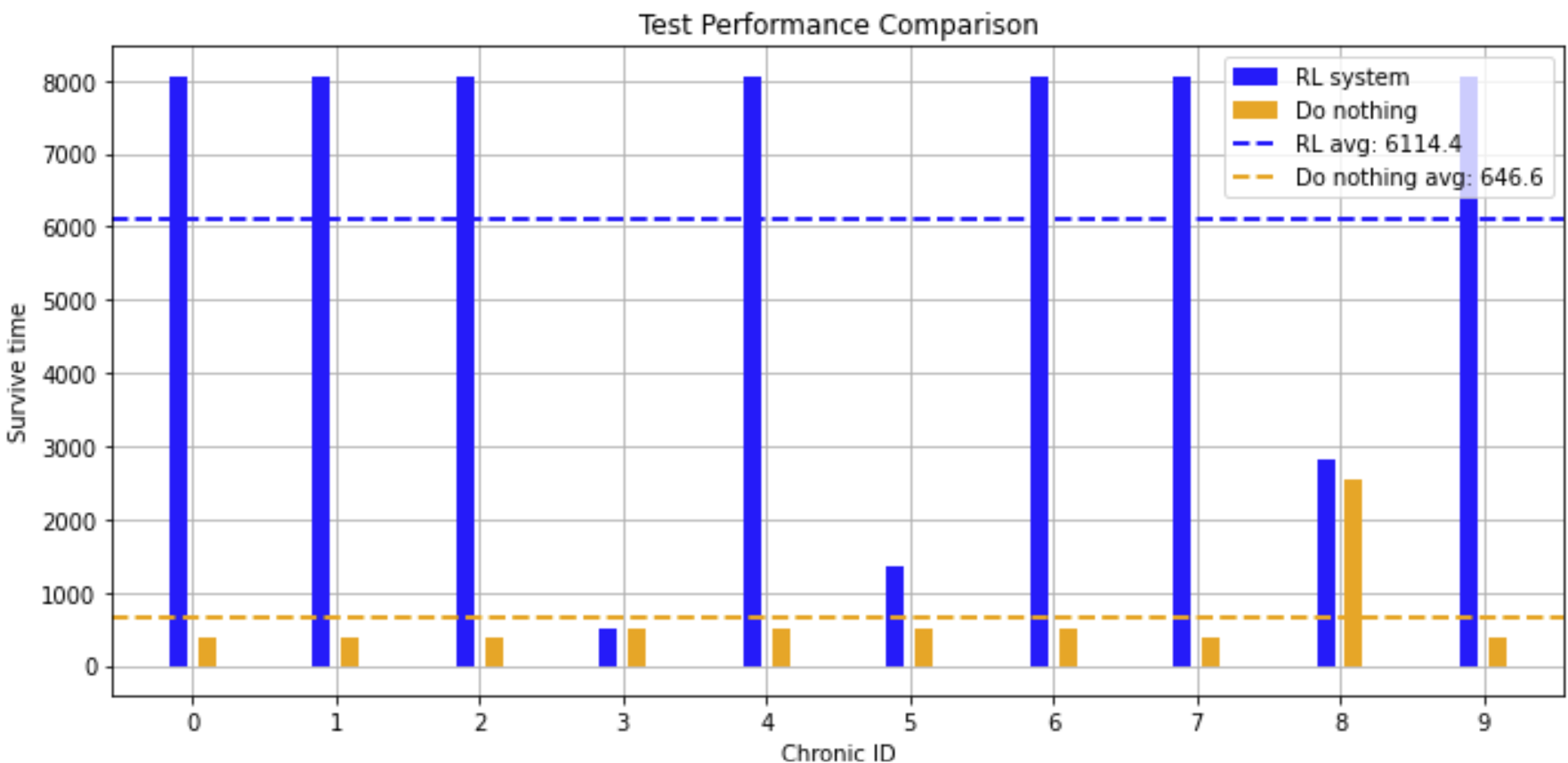}
    \caption{Comparison of the performance of our RL agent against the Do-Nothing baseline over the validation set (top) and the test set (bottom). }
    \label{fig: performances}
\end{figure}

As shown in Fig. \ref{fig: performances}, the proposed architecture outperforms the \textit{Do-Nothing} baseline on both validation and test sets. The $x$-axis denotes the id of the specific chronic (10 in total for both validation and test sets), while the $y$-axis reports the survive time.

For a deeper understanding of the model's components, an ablation study was conducted. The contribution of the components of the proposed architecture was analyzed by removing them one at a time and observing the impact on performance. A summary of the results is provided in Tab. \ref{tab:ablation}.

The ablation study highlights that components such as DQfD and the GNN are essential to achieve competitive performance within a limited amount of time. In contrast, the impact of bootstrapped reward shaping is less evident in the current environment. However, it is reasonable to expect that in larger and more complex grid topologies, where reward signals may be highly diluted, reward shaping could play a more critical role in stabilizing and accelerating the learning process.

\begin{table}[t!]
\centering
\caption{Ablation study results: average survival time on validation and test sets. The best result per column is in bold. The first row (Do-Nothing) reports the performance of the baseline.}
\label{tab:ablation}
\renewcommand{\arraystretch}{1.1}
\begin{tabular}{p{4cm}p{3cm}p{3cm}}
\toprule
\textbf{Configuration} & \textbf{Validation Perf.} & \textbf{Test Perf.} \\
\midrule
Do-Nothing (baseline)   & 997.7 & 646.6 \\
\midrule
No DQfD                 & 1985 & 1878  \\
No GNN                  & 1206.4 & 785.2 \\
No Reward Shaping       & \textbf{5667.1} & 5324.3 \\
Complete System         & 5452.2 & \textbf{6114.4} \\
\bottomrule
\end{tabular}
\end{table}

\begin{figure}[t!]
    \centering
    \includegraphics[width=0.55\textwidth]{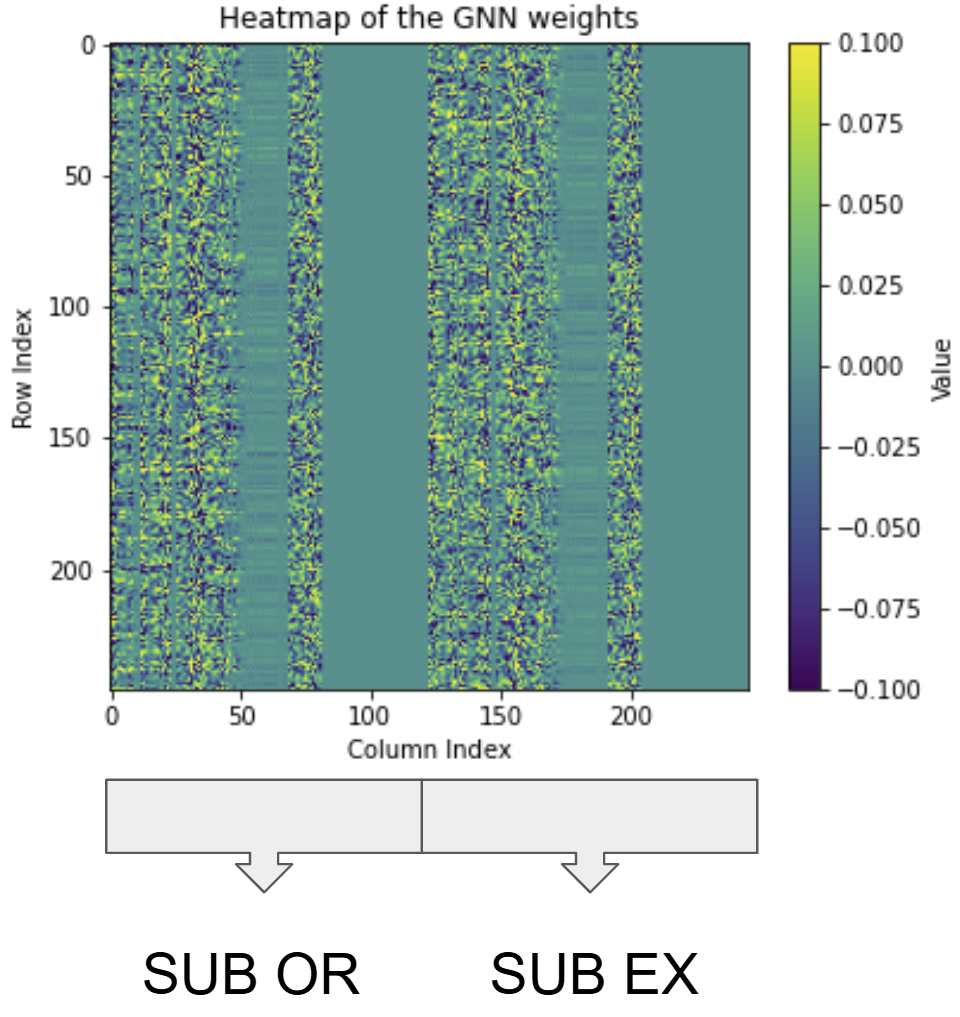}
    \caption{Heatmap of the weights in the first GNN layer.}
    \label{fig:gnn_heatmap}
\end{figure}

Finally, the learned weights of the first GNN layer are analyzed to provide insights about the generalization capabilities of the model.
Fig.~\ref{fig:gnn_heatmap} displays a heatmap of these weights, providing a qualitative perspective on how the model processes structured power grid information.

\begin{table}[t!]
\centering
\caption{Inference Time Comparison: Proposed approach against Expert agent.}
\renewcommand{\arraystretch}{1.1}
\label{tab:inference_time}
\begin{tabular}{p{2.7cm}p{5cm}}
\toprule
\textbf{Model} & \textbf{Inference Time (avg ± std) [s]} \\
\midrule
Proposed Model & 0.187 ± 0.145 \\
Expert Agent & 2.56 ± 0.223 \\
\bottomrule
\end{tabular}
\end{table}

Recall that the input vector processed by the GNN is formed by concatenating the feature vectors of the origin and extremity substations of each power line. This structural symmetry is reflected in the heatmap, which displays a clearly symmetrical pattern with respect to the column indices corresponding to these two segments. This symmetry suggests that the GNN has effectively learned the structure of its input representation, potentially indicating strong generalization capabilities.

\paragraph{Computational Complexity.}
To analyze the efficiency of the model, the computational requirements of the proposed approach were investigated. In general, simulating a power grid is, by nature, a computationally intensive task. The plain power grid simulation takes on average 0.1097 seconds to simulate a single step. Thus, it takes almost 15 minutes to simulate all the 8064 time steps of an episode. The inference time of the proposed approach is compared against the one of the Expert agent used to collect expert demonstrations. The results, provided in Tab. \ref{tab:inference_time}, reveal the computational efficiency of the proposed model compared to the one of the simulation-based expert agent.

\section{Related Works}
\label{sec:relatedworks}

\textit{Distributed RL}. Most of the RL methods developed to control power grids are related to the L2RPN competitions \citep{kelly2020reinforcement, serre2022reinforcement}. An overview of the solutions proposed during the first edition is reported in \citep{marot2021learningrunpowernetwork}. The winning solutions of the last two editions are presented in \citep{dorfer2022power} and \citep{artelys2023}, respectively. All the solutions proposed so far severely restrict the action space and are based on a single agent or multiple agents acting on the entire grid.

The first approach that introduced multi-agent solutions within the power grid topology management problem was presented in \citep{manczak2023hierarchicalreinforcementlearningpower}. The solution is a two-layer distributed RL model: a single RL-manager and multiple low-level agents (one for substation). When it is time to play, the manager selects the low-level agent that consequently plays on its controlled substation.
Another Multi-Agent approach for the power grid topology management problem was implemented by \citep{vandersar2023multiagentreinforcementlearningpower}. In this work, each substation is managed by a different agent trained with PPO~\citep{schulman2017proximalpolicyoptimizationalgorithms}. The high-level logic deciding which agent comes to play upon a critical situation is a rule-based logic called CAPA, which was first introduced by \citep{yoon2021winning} but for a different scope. When a line is in overload, the CAPA logic selects as candidate agents the two substations extremities to the overloaded power line, and later, the substation with the highest average line utilization is selected among the two candidates.
The CAPA logic presents an important limitation in the performance of such a distributed architecture: it becomes difficult for the agents to develop long-term cooperative strategies. The two previous works have been joined together in the recent work \citep{demol2025centrallycoordinatedmultiagentreinforcement}. Here, the authors compare several two-layer RL solutions. In particular, the effect of CAPA high-level logic against an RL manager. The results show that the RL manager outperforms the system with a CAPA high-level manager in large power grids.

This work aims to go one step further by addressing both the observation- and action-space dimensions of the scalability challenge. We propose a distributed framework in which every agent can observe just a limited part of the state space and take only a small number of actions, but all the agents cooperate to achieve a common goal. 

\noindent\textit{Graph Neural Networks.} There is no well-established and standard framework to use GNNs within the context of power grid management via topology changes. Many approaches have been proposed over the years, differing in many aspects of the procedures, even in the representation of the power grid as a graph. A comprehensive survey about the use of GNN within power grids is provided in \citep{hassouna2024graphreinforcementlearningpower}.
The first work to introduce GNNs for power grid topology control was \cite{9347305}, where the GNN was used as a feature extractor for a standard DQN agent. This solution represented the power grid as a heterogeneous graph in which the nodes were generators, loads and ends of power lines, with two elements linked if connected to the same bus. 
The same graph representation was adopted in several subsequent works \citep{9830198,9535409}.
The first solution, introducing a new graph representation, is the winning solution of the L2RPN 2020 challenge \citep{yoon2021winning}. It represented the power grid as a graph in which nodes are substations, thus using a homogeneous graph representation.
Notably, a heterogeneous graph representation of the power grid raises two important problems: the so-called “bus-bar information asymmetry” and the need for heterogeneous GNN models. The former highlights the fact that, within heterogeneous graph representation, elements linked to different buses on the same substation are not connected together. Therefore, information about potential connection is lost. The latter, instead, underlines a common problem of using standard GNNs with heterogeneous graphs: they fail to generalize.
This topic has been analyzed by \citet{dejong2025generalizablegraphneuralnetworks} that solved both problems by slightly improving the heterogeneous graph representation and by using a heterogeneous GNN model. However, the adoption of a heterogeneous GNN inevitably increases the model's complexity.

\section{Conclusions}
\label{sec:conclusions}
In this work, we introduced a novel distributed graph-based RL algorithm to perform power grid operation control. The method consists of a network of distributed low-level agents, each controlling a different power line, coordinated by a high-level manager. Each low-level agent operates solely on its associated power line and receives a local observation preprocessed by a shared graph neural network. This procedure makes it possible to decompose the action space and the observation space simultaneously, enabling a fully scalable solution. To further improve learning efficiency and stability, the framework incorporates Deep Q-Learning from Demonstrations and potential-based reward shaping techniques. The results obtained in the Grid2Op simulation environment demonstrate the effectiveness of the proposed graph-based framework. The method significantly outperforms the standard Do-Nothing baseline in terms of survival time, while also showing much lower inference time compared to the Expert System. These outcomes confirm the scalability and efficiency of the architecture, particularly in scenarios requiring rapid decision-making across large-scale networks.

The main contribution of this work lies in the novel application of Graph Neural Networks to decompose the observation space while constructing informative local observations for the decentralized agents. Both the conversion of the environment’s observation into a homogeneous graph structure and the use of a shared Graph Neural Network at the base of all distributed agents are carefully motivated design choices. Together, they contribute significantly to the strength and coherence of the proposed approach, enabling structured local observations and efficient information sharing. The combination of Deep Q-Learning from Demonstrations and potential-based reward shaping also plays an important role in the proposed framework. Deep Q-Learning from Demonstrations proved to be essential for reaching competitive performance within a limited amount of time, while potential-based reward shaping appears to be a promising research direction for addressing the problem of diluted rewards in large-scale power networks.

\textit{Limitations and Future Works.}
Despite its strengths, the proposed approach still presents some limitations. Firstly, although the low-level agents receive local observations, the manager needs access to a complete global view of the environment to make decisions. This global view limits the scalability of the overall architecture. To overcome this limitation, multi-layer architectures represent a promising research direction. In these architectures, multiple high-level managers control different, independent portions of the power grid, identified through power grid decomposition methods.
Secondly, applying imitation learning assumes the availability of a dataset of expert demonstrations, which is not always the case. This is particularly problematic when dealing with large, unmanageable power grids for which expert strategies are unknown. Lastly, while the model effectively decomposes the action space, it still relies on a single-step simulation at runtime to perform a greedy action-space reduction. This step likely helps achieve good performance in a shorter amount of time, but it may not be strictly necessary. Alternative action space reduction techniques should be explored. A promising direction is the $N$-$1$ criterion proposed by \citep{demol2025centrallycoordinatedmultiagentreinforcement, vandersar2025optimizingpowergridtopologies}. This criterion statically reduces the action space by nearly half by retaining only those actions that preserve $N$-$1$ stability, ensuring that the power grid remains operational even if a random power line is disconnected.

Future developments should focus on addressing some of the mentioned limitations. The development of multi-layer solutions, with multiple region-based high-level managers coordinated by a top-level rule-based logic, represents a possible approach to avoid using complete global views of the environment. Additionally, alternative action space reduction techniques, as the previously mentioned $N$-$1$ criterion, should be explored in order to significantly reduce the inference time.  Lastly, an interesting research direction, introduced by the use of a single shared GNN, is GNN-based transfer learning. This involves using a GNN trained on a smaller power grid as initialization for a model operating on a larger grid. This approach could significantly accelerate convergence in large-scale scenarios.

\vspace{0.1cm}

\noindent\textbf{Acknowledgments.} The authors acknowledge the project AI4REALNET that has received funding from European Union’s Horizon Europe Research and Innovation programme under the Grant Agreement No 101119527. Views and opinions expressed are, however, those of the authors only and do not necessarily reflect those of the European Union. Neither the European Union nor the granting authority can be held responsible for them.

\small
\bibliographystyle{unsrtnat}
\bibliography{biblio}

\end{document}